\documentclass{article}
\usepackage{graphicx}

\usepackage{times}
\usepackage{latexsym}

\usepackage{blindtext}
\usepackage{CJKutf8}
\usepackage{latexsym}
\usepackage{mathrsfs}
\usepackage{amsfonts}
\usepackage{amssymb}
\usepackage{amsmath}
\usepackage{bm}
\usepackage{url}
\usepackage{hyperref}
\usepackage{times}
\usepackage{graphicx} 
\usepackage{subfigure}
\usepackage{multicol}
\usepackage{multirow}
\usepackage{booktabs}
\usepackage{colortbl}
\usepackage[normalem]{ulem} 
\newcommand\hl{\bgroup\markoverwith
  {\textcolor{yellow}{\rule[-.5ex]{2pt}{2.5ex}}}\ULon}
  
\usepackage{microtype}
\usepackage{xspace}

\newcommand{\method}{CNewSum\xspace}

\begin{document}
\title{\method: A Large-scale Chinese News Summarization Dataset with Human-annotated Adequacy and Deducibility Level}

\author{Danqing Wang, Jiaze Chen, Xianze Wu, Hao Zhou and Lei Li\thanks{Work is done while the corresponding author was at ByteDance.} \\
\{wangdanqing.122,chenjiaze,zhouhao.nlp\}@bytedance.com \\
*lilei@cs.ucsb.edu}
\date{June, 2021}

\maketitle              

\begin{abstract}
Automatic text summarization aims to produce a brief but crucial summary for the input documents.
Both extractive and abstractive methods have witnessed great success in English datasets in recent years. 
However, there has been a minimal exploration of text summarization in Chinese, limited by the lack of large-scale datasets.
In this paper, we present a large-scale Chinese news summarization dataset \method, which consists of 304,307 documents and human-written summaries for the news feed. It has long documents with high-abstractive summaries, which can encourage document-level understanding and generation for current summarization models.
An additional distinguishing feature of \method is that its test set contains adequacy and deducibility annotations for the summaries. 
The adequacy level measures the degree of summary information covered by the document, and the deducibility indicates the reasoning ability the model needs to generate the summary.
These annotations can help researchers analyze and target their model performance bottleneck.
We examine recent methods on \method and release our dataset\footnote{It is available at  \url{https://dqwang122.github.io/projects/CNewSum/}.} to provide a solid testbed for automatic Chinese summarization research.



\end{abstract}

\section{Introduction}
\label{sec:intro}
\begin{CJK*}{UTF8}{gkai}
\begin{table}[ht]\footnotesize
  \centering
  \caption{An example of our \method dataset. `Sentence Label' is the id of sentences selected as the supervised signals for extractive models via the greedy algorithm. All information of the summary can be found in the document, so its adequacy and deducibility level are 1.}
  \label{tab:example}%
    \begin{tabular}{l}
    \toprule
    \multicolumn{1}{p{12cm}}
    {\textbf{Article}  [0]图为在广元市朝天区发现的白耳夜鹭。[1]广林局供。[2]中新网广元3月15日电。[3]记者15日从四川省广元市野生动物救治中心获悉：近日，该市朝天区东溪河乡的群众发现一只受伤的``怪鸟”引起多方关注，随后，上报到广元市林业部门。[4]后经当地野生动物保护专家鉴定“怪鸟”为世界最濒危鸟类白耳夜鹭。…… [7]该鸟是我国特有的珍稀鸟类、国家二级保护动物白耳夜鹭，被列为世界最濒危的30种鸟类之一，目前全世界仅存1000余只 …… [14]此后一直再没有关于该鸟踪迹的报道。
    \newline{}(\textit{[0]The picture shows the Gorsachius magnificus in Chaotian District of Guangyuan City. [1]Supplied by Guangyuan Forestry Department. [2]Xinhua News Agency, Guangyuan, March 15. [3]Reporters learned from the Wildlife Treatment Center of Guangyuan City, Sichuan Province on the 15th that recently, the discovery of an injured "strange bird" by the local people in Dongxihe village, Chaotian District of the city attracted much attention and was subsequently reported to the forestry department of Guangyuan City. [4]After that, local wildlife protection experts identified the "strange bird" as the world's most endangered bird, the Gorsachius magnificus.......[7]It is a rare bird unique to our country and a national second-class protected animal, the Gorsachius magnificus. It has been listed as one of the world's most endangered 30 species of birds. At present, there are only about 1,000 birds in the world......[14]There have been no reports of the bird’s trace since then.})} \\
    \midrule
    \multicolumn{1}{p{12cm}}
    {\textbf{Summary}  今日获悉，广元一市民发现受``怪鸟”，经鉴定系世界濒危鸟类白耳夜鹭，全球仅存1000只。\newline{}(\textit{It was reported today that a citizen of Guangyuan found an injured ``strange bird", which was identified as a world-endangered bird, the white-eared night heron, of which only 1,000 exist worldwide.})} \\
    \midrule
    \textbf{Sentence Label}: \quad \{0,4\} \quad \textbf{Adequacy Level}: 1 \quad \textbf{Deducibility Level}: 1 \\
    \bottomrule
    \end{tabular}%
\end{table}%
\end{CJK*}

Text summarization is an important task in natural language processing, which requires the system to understand the long document and generate a short text to summarize its main idea.
There are two primary methods to generate summaries: \textit{extractive} and \textit{abstractive} methodology. Extractive methods select semantic units from the source document and reorganize them into a consistent summary, while abstractive models generate summaries using words and phrases freely. 
Benefiting from pre-trained language models~\cite{devlin2019bert,lewis2020bart,liu2019roberta}, much progress has been made on English summarization datasets, such as Newsroom~\cite{grusky2018newsroom}, CNN/DailyMail~\cite{hermann2015teaching}, and NYT~\cite{sandhaus2008new}.

However, the lack of high-quality datasets in other languages, such as Chinese, limits further researches on summarization under different language habits and cultural customs. 
Currently, most Chinese summarization datasets are collected from Chinese social media Weibo, which are limited to a 140-character length~\cite{gao2019abstractive,hu2015lcsts}.
Some other datasets are scraped from news websites, such as Toutiao~\cite{hua2017overview} and ThePaper~\cite{liu2020clts}. However, those datasets are either small-scale or of low quality.


In this paper, we present a large-scale Chinese news summarization dataset, \method, to make up for the lack of Chinese document-level summarization, which can become an important supplement to current Chinese understanding and generation tasks. Different from previous summarization datasets crawled from news websites, we called for news articles from hundreds of thousands of press publishers and hired a team of expert editors to provide human-written summaries for the daily news feed. During the summarization process, the editors may perform simple reasoning or add external knowledge to make the summary more reader-friendly.
Thus, we further investigate our test set and explore how much knowledge the models need to generate a human-like summary. Specifically, we ask annotators to determine two questions: 1) \textbf{Adequacy}: \textit{Is the information of summaries self-contained in the source document?} 2) \textbf{Deducibility}: \textit{Can the information be deduced from the source document directly, or needs external knowledge?} We provide these two scores for each example in the test set.
Table \ref{tab:example} is an example of our dataset.

Our main contributions are as follows: 

(1) We propose a large-scale Chinese news summarization dataset collected from hundreds of thousands of news publishers. We hire a team of expert editors to write summaries for the news feed.


(2) In order to figure out how much knowledge the model needs to generate a human-like summary, we manually annotate the adequacy and deducibility scores for our test set.

(3) We also provide several extractive and abstractive baselines, which makes the dataset easy to use as the benchmark for Chinese summarization tasks. 

\section{Related work}
\label{sec:related}

\paragraph{News Summarization Dataset}

Most news summarization datasets focus on English, and here we give a brief introduction to some popular ones and list the detailed information in the first part of Table \ref{tab:dataset}. 
NYT is a news summarization dataset constructed from New York Times Annotated Corpus~\cite{sandhaus2008new}. We tokenize and convert all text to lower-case, follow the split of Paulus et al.~\cite{paulus2018deep}. 
The CNN/DailyMail question answering dataset~\cite{hermann2015teaching} modified by Nallapati et al.~\cite{nallapati2017summarunner} and See et al.~\cite{see2017get} is the most commonly-used dataset for single-document summarization. It consists of online news articles with several highlights. Those highlights are concatenated as the summary.
Newsroom~\cite{grusky2018newsroom} is a large-scale news dataset scraped from 38 major news publications, ranging from business to sports. These summaries are often provided by editors and journalists for social distribution and search results. 

\paragraph{Chinese Summarization Dataset}
There are also several Chinese summarization datasets in other domains~\cite{gao2019how,huang2020generating,xi2020global}, but here we only discuss news summarization datasets. The detailed statistics are listed in the second part of Table \ref{tab:dataset}.
The LCSTS~\cite{hu2015lcsts} is a large-scale Chinese social media summarization dataset. It is split into three parts, and part II and part III are usually used as development and test set after filtering out low-quality examples.
RASG~\cite{gao2019abstractive} collects the document-summary-comments pair data for their reader-aware abstractive summary generation task. It utilizes users' comments to benefit the generation of the abstractive summary of main content. The document is relatively short and has about 9 comments as a complement. 
TTNews~\cite{hua2017overview} is provided for NLPCC Single Document Summarization competition\footnote{http://tcci.ccf.org.cn/conference/2018/taskdata.php}, including 50,000 training examples with summaries and 50,000 without summaries.
CLTS~\cite{liu2020clts} is a Chinese summarization dataset extracted from the news website ThePaper. It contains more than 180,000 long articles and summaries written by editors of the website.


\section{The \method Dataset}
\label{sec:dataset}
\subsection{Data Collection}

We receive news submissions from hundreds of thousands of press publishers\footnote{The press publishers include thepaper.cn, wallstreetcn.com, cankaoxiaoxi.com, yicai.com, and so on. They submit their articles in web format to our company. These publishers retain any copyright they may have in their content and grant us a royalty-free, perpetual license to use, copy, edit and publish their content.}. These articles do not have corresponding summaries, so we hire a team of expert editors to provide human-written summaries for the daily news feed. Each example will be double-checked by different experts to ensure its quality. 
We construct \method by extracting news articles from 2015 to 2020\footnote{These data have been checked for legality and can be released for research use.}
and filtering summaries with less than 5 words. We further limit the length of documents to 50-5000. 





Finally, we obtain a Chinese news corpus with 304,307 document-summary pairs. It is split into training/validation/test by 0.9/0.05/0.05. Besides, we compare document sentences with human-written summaries and use the greedy algorithm following~\cite{nallapati2017summarunner} to get the \textsc{Oracle} sentences with label 1 as the signals for extractive summarization. 

\begin{table}[htbp]\footnotesize\setlength{\tabcolsep}{3pt}
  \centering
  \caption{The summarization datasets. The top part contains the commonly-used English news summarization and the bottom contains the Chinese summarization datasets. `-' means the original dataset does not provide the standard spit for train/dev/test set. For TTNews, we only take training examples with summaries into consideration. `*' includes 2,000 evaluation examples for NLPCC2017 and 2,000 for NLPCC2018.}
  \label{tab:dataset}%
    \begin{tabular}{lrrr|r|rc|c}
    \toprule
    Dataet & \multicolumn{1}{c}{\textbf{Train}} & \multicolumn{1}{c}{\textbf{Dev}} & \multicolumn{1}{c|}{\textbf{Test}} & \multicolumn{1}{c|}{\textbf{Total}} & \multicolumn{1}{c}{\textbf{Article}} & \multicolumn{1}{c|}{\textbf{Summary}} & \multicolumn{1}{c}{\textbf{Source}} \\
    \midrule
    NYT~\cite{sandhaus2008new}   & 589,282  &        32,737  &        32,739  &     654,758  & 552.14 & 42.77 & New York Times\\
    CNNDM~\cite{hermann2015teaching} &     287,227  &        13,368  &        11,490  &     312,085  & 791.67 & 55.17 & CNN \& Daily Mail\\
    Newsroom~\cite{grusky2018newsroom} &     995,041  &      108,837  &      108,862  &   1,212,740  & 765.59 & 30.22 & 38 news sites \\
    \midrule
    LCSTS~\cite{hu2015lcsts} & 2,400,591 & 8,685 & 725   &   2,410,001  & 103.7 & 17.90  & Weibo \\
    RASG~\cite{gao2019abstractive}  & 863,826 & -     & -     &      863,826  & 67.08 & 16.61 & Weibo \\
    TTNews~\cite{hua2017overview} & 50,000 & - & 4,000* & 54,000  & 747.20 & 36.92 & Toutiao \\
    CLTS~\cite{liu2020clts} & 148,317 & 20,393 & 16,687 & 185,397 & 1363.69 & 58.12 & ThePaper \\
    \method & 275,596 & 14,356 & 14,355 & 304,307  & 790.55 & 37.58 &  News publishers \\
    \bottomrule
    \end{tabular}%
\end{table}%

\subsection{Adequacy and Deducibiltiy Annotation}

Analyzing our dataset, we find that the expert editors often perform some reasoning or add external knowledge to make the summary more friendly for the readers. For example, a precise figure (2,250) may be summarized as an approximate number (more than 2000). In another case, a specific date will be converted to a relative time based on the time of publication, e.g., tomorrow. This information is not directly available in the original document. Thus, we wonder how much knowledge the model needs to generate the human-like summary. Inspired by~\cite{chen2016thorough}, we ask annotators to answer the two questions for each document-summary pair in our test set: 

1) \textbf{Adequacy} \textit{Does necessary information of the summary has been included in the document?} For example, all words in the summary can be directly found in the document, or they have synonyms or detailed descriptions in the original text. Under these circumstances, the summary is labeled as 1. Otherwise, the summary is labeled as 0.

2) \textbf{Deducibility} \textit{Can the information of the summary be easily inferred from the document?} Unit conversion, number calculation, and name abbreviations that can be inferred are labeled as 1. In contrast, complex conclusions with no direct mentions in the original document are labeled as 0. 

For each question, the annotators should choose 0 or 1. We hired a team of 12 employees to annotate the test set\footnote{We paid 1 RMB (0.15 dollars) for each example, and the average hourly wage is 60 RMB (the minimum hourly wage is 24 RMB).}. We first trained these employees on basic annotation rules, and they were required to annotate 100 examples and then be checked and corrected by us. Two expert annotators were employed to control quality. They were asked to sample 10\% examples from each annotator and recheck the annotation. If one's consistent rate is less than 95\%, all annotations of this annotator will be returned and re-annotated. An example is consistent only if the two experts and the annotator agree on their answers; otherwise, the example will be further discussed.

\begin{table}[htbp]\footnotesize\setlength{\tabcolsep}{2pt}
  \centering
  \caption{The statistics of news summarization datasets. \textit{Coverage}, \textit{Density} and \textit{Compression} are introduced by~\cite{grusky2018newsroom}. The Bigram, Trigram and 4-gram are the n-gram novelty (\%). The novelties of NYT/CNNDM/Newsroom are from~\cite{narayan2019what}. For Chinese data, it is calculated by words.}
  \label{tab:anlysis}%
    \begin{tabular}{lcccccc}
    \toprule
    \textbf{Dataset} & \multicolumn{1}{c}{\textbf{Coverage$\downarrow$}} & \multicolumn{1}{c}{\textbf{Density$\downarrow$}} & \multicolumn{1}{c}{\textbf{Compression$\uparrow$}} & \multicolumn{1}{c}{\textbf{Bigram$\uparrow$}} & \multicolumn{1}{c}{\textbf{Trigram$\uparrow$}} & \multicolumn{1}{c}{\textbf{4-gram$\uparrow$}} \\
    \midrule
    \multicolumn{1}{l}{NYT} & 0.83  & 3.50  & 24.19 & 55.59 & 71.93 & 80.16 \\
    CNNDM & 0.85  & 3.70  & 13.76 & 49.70 & 70.20 & 79.99 \\
    Newsroom & 0.82  & 9.50  & 36.03 & 46.80 & 58.06 & 62.72 \\
    \midrule
    LCSTS & 0.54 & 1.23 & 6.61 & 80.29 & 90.92 & 94.53 \\
    RASG & 0.61 & 2.52 & 7.27 & 67.89 & 76.94 & 80.15 \\
    TTNews & 0.76 & 3.21 & 22.24 & 61.09 & 76.30 & 83.64 \\
    CLTS & 0.99 & 28.73 & 24.81 & 5.14 & 8.08 & 10.36 \\
    \method & 0.76  & 2.77  & 20.83 & 63.29 & 78.54 & 85.64 \\
    \bottomrule
    \end{tabular}%
\end{table}%

\subsection{Dataset Analysis}
\label{sec:analysis}
As shown in Table \ref{tab:dataset}, our \method dataset has a similar scale with the most popular English summarization dataset CNNDM, which is suitable for training and evaluating different summarization models. For the Chinese dataset, the average length of the document and the summary are significantly longer than datasets collected from Weibo and similar to TTNews. 

Following Grusky et al.~\cite{grusky2018newsroom}, we also use \textit{Coverage}, \textit{Density} and \textit{Compression} to characterize our summarization dataset. \textit{Coverage} measures the overlap degree of the extractive fragment between the article and summary, and \textit{Density} measures the average length of the extractive fragment. \textit{Compression} is the ratio of the article length to the summary length. In addition, we calculate the n-gram novelty of the summary, which is the percentage of n-grams that do not appear in the document, as described in~\cite{narayan2019what}. 
The results are shown in Table \ref{tab:anlysis}. We can find that the datasets collected from Weibo usually have lower coverage and density ratio, with high compression and novelty. This indicates that the summaries for these short documents are more abstractive. For news article summarization, CLTS copies most words of the summary from the document directly, which is indicated by the highest coverage, density and the lowest novelty. Our \method provides a large-scale document-level summarization dataset with comparable abstractiveness with short social media datasets.




Since all adequacy summaries can be inferred from the document, the A=1 \& D=0 is meaningless. For the summarization models, the examples with A=1 \& D=1 are relatively easy to generate, and the examples with A=0 \& D=1 ask for some inference abilities. The A=0 \& D=0 cannot be solved with the original document and may need the help of external knowledge.

We find that more than 91.08\% examples are adequate and deducible, but the rest lack essential information. For the remaining 4.11\% examples with $D = 1$, the information can be inferred from the document. Typically, ``2005-2015" will be summarized as ``ten years" which requires the model to do simple calculations. The rest summaries are factual but need external knowledge. News articles from the websites are time-sensitive and are filled with pictures. The editors often write the summary based on the time of the event and the image, which will cause the relative time, such as `yesterday', and the picture description to appear in the summary. In addition, famous people will be mapped to their position in the summary, such as Obama and the American president of that time. It is difficult for the model to deduce such information from the news text without additional information. We keep these in our dataset to simulate real-world data distribution and let researchers evaluate the model performance from different aspects.

\section{Experiment}
\label{sec:exper}

We train several summarization models on our \method. These systems include both abstractive and extractive methods, and the performance can serve as the baseline for future work.

\subsection{Models}

\begin{table}[htbp]
  \centering
  \caption{Results on the test set of \method. The first part contains the Lead and Oracle baseline. The second and third part are extractive and abstractive summarization models.}
  \label{tab:results}%
    \begin{tabular}{lccc}
    \toprule
    \textbf{Models} & \multicolumn{1}{c}{\textbf{ROUGE-1}} & \multicolumn{1}{c}{\textbf{ROUGE-2}} & \multicolumn{1}{c}{\textbf{ROUGE-L}} \\
    \midrule
    \textsc{Lead}  & 30.43 & 17.26 & 25.33 \\
    \textsc{Oracle} & 46.84 & 30.54 & 40.08 \\
    \midrule
    TextRank~\cite{mihalcea2004textrank} & 24.04 & 13.70 & 20.08 \\
    NeuSum~\cite{zhou2018neural} & 30.61 & 17.36 & 25.66 \\
    Transformer-ext & 32.87 & 18.85 & 27.59 \\
    BERT-ext & 34.78 & 20.33 & 29.34 \\
    \midrule
    Pointer Generator~\cite{see2017get}     & 25.70 & 11.05 & 19.62 \\
    Transformer-abs & 37.36 & 18.62 & 30.62 \\
    BERT-abs & \textbf{44.18} & \textbf{27.37} & \textbf{38.32} \\
    \bottomrule
    \end{tabular}%
\end{table}%

\paragraph{Baseline}
We calculate two popular summarization baselines for our dataset. \textsc{Lead} is a common lower bound for news summarization dataset~\cite{grusky2018newsroom,nallapati2017summarunner,see2017get}, which selects the first several sentences as the summary. Here, we choose the first two sentences.
For \textsc{Oracle}, we concatenate the sentences with label 1 with their original order in the document.

\paragraph{Extractive Models}
TextRank~\cite{mihalcea2004textrank} is a simple unsupervised graph-based extractive method. It takes sentences as nodes and calculates the node importance based on eigenvector centrality. 
NeuSum~\cite{zhou2018neural} jointly scores and selects sentences for extractive summarization. Transformer~\cite{vaswani2017attention} is a well-known sequence-to-sequence model based on the self-attention mechanism, the pre-trained language models such as BERT~\cite{devlin2019bert} trained on large corpus\footnote{Since the bert-base-chinese model of Google does not perform well in our dataset, we train a Chinese BERT language model with Chinese news articles.} have shown great performance. We use the code\footnote{https://github.com/nlpyang/PreSumm} provided by BERTSum~\cite{liu2019text} and follow the experimental settings to apply the Transformer and BERT to extractive summarization, which are named Transformer-ext and BERT-ext. Both of them use a 6-layer Transformer with hidden size 768 and feed-forward filter size 2048 as the document encoder. The sigmoid layer is put on the top to score the sentences. We choose the top sentence as the summary due to the average sentence number (1.03) of the ground truth summary. 


\paragraph{Abstractive Models}
Pointer Generator~\cite{see2017get} is the pointer-generator network which is a commonly-used encoder-decoder abstractive summarization model with the copy and coverage mechanism. We also use the Transformer encoder and decoder for abstractive summarization. They are called Transformer-abs and BERT-abs to distinguish them from the above extractive models. These Transformer-based abstractive models use the same transformer encoder as the extractive ones and a transformer decoder with 6 layers for generation.

\begin{table}
  \centering
  \caption{The results of models on different adequacy and deducibility level.}
  \label{tab:adres}%
    \begin{tabular}{clccc}
    \toprule
    \textbf{Model} & \multicolumn{1}{c}{\textbf{Category}} & \multicolumn{1}{c}{\textbf{ROUGE-1}} & \multicolumn{1}{c}{\textbf{ROUGE-2}} & \multicolumn{1}{c}{\textbf{ROUGE-L}} \\
    \midrule
    \multirow{3}[2]{*}{Transformer-ext} & A=1\&D=1 & 33.16 & 19.19 & 27.88 \\
      & A=0\&D=1 & 30.89 & 15.60 & 25.38 \\
      & A=0\&D=0 & 28.92 & 14.88 & 23.74 \\
    \midrule
    \multirow{3}[2]{*}{Transformer-abs} & A=1\&D=1 & 37.54 & 18.85 & 30.83 \\
      & A=0\&D=1 & 36.36 & 16.70 & 29.63 \\
      & A=0\&D=0 & 34.73 & 15.95 & 27.52 \\
    \midrule
    \multirow{3}[2]{*}{BERT-ext} & A=1\&D=1 & 35.05 & 20.67 & 29.62 \\
      & A=0\&D=1 & 32.81 & 16.90 & 27.05 \\
      & A=0\&D=0 & 31.07 & 16.57 & 25.72 \\
    \midrule
    \multirow{3}[2]{*}{BERT-abs} & A=1\&D=1 & 44.51 & 27.76 & 38.70 \\
      & A=0\&D=1 & 41.75 & 23.64 & 35.34 \\
      & A=0\&D=0 & 40.18 & 23.34 & 33.60 \\
    \bottomrule
    \end{tabular}%
\end{table}%


\begin{CJK*}{UTF8}{gkai}
\begin{table}\footnotesize\setlength{\tabcolsep}{1pt}
  \centering
  \caption{An example for abstractive summarization models. The text with underline is directly copied from the original article, and the text with wavy underline contains fake information. }
  \label{tab:case}%
    \begin{tabular}{lp{10cm}}
    \toprule
    \textbf{Article} & 英雄联盟神秘预告再现。官方最新\underline{发布了一个短片视频}，其短片的名称是“他已归来”。而最近更新的巨神峰新故事中就有描述星灵的，难道新英雄是星灵来自银河？今日，国外的LOL官方社交媒体上，放出了一个预告短片，名称为“他已归来”。\underline{短片内容为，潘森正在凝视夜空中被星云所围绕的亮光}。有人猜测，视频中的场景为潘森故事《巨神之枪》中的末尾内容，也是巨神峰新故事中所描述的《星灵》。歪果仁点评：Gigathor：天啊，下一个新英雄是银河系的！MrBananaHump：跟你们开玩笑呐，这只不过是巴德。SoSaysCory：应该是潘森的兄弟，潘林将会加入峡谷，技能与潘森一样，他们将会成为有史以来最强力的下路组合。Sharjo：将会有全新的巨神峰英雄了！潘森新的背景故事已提到了这个，在《巨神之枪》故事的结尾，指出了新的星灵到来。来自另一个次元的潘森老朋友将会和我们见面了！太酷了！DracCusS：感觉是：a)新英雄。b)潘森模型更新。c)宝石重做？ 
    
    \textit{League of Legends released a mysterious trailer and the official latest \underline {posted a short video}. The name of the short film is ``He Has Returned". In the recent new story of Mount Titan, there is a description of the Protoss. Will the new hero be the Protos from the Milky Way? Today, a short trailer was released on the official social media of LOL abroad, titled ``He Has Returned." \underline{The content of the video is, Pan Sen stares at the bright light surrounded by} \underline{nebula in the night sky}. Some people guess, the scene in the video is the content of Pan Sen's story ``Gun of the Titans", and the new story of Mount Titan is also described in the ``Protoss". Foreigns Review: Giant: Oh my god, the next new hero is from the galaxy! BananaHump: I'm joking with you, this happens to be Bud. SoSaysCory: It should be Pan Sen's brother. Pan Lin will join the canyon. The skills are the same as Pan Sen. They will become the most concentrated bot combination ever. Sharjo: There will be a brand new Giant Mountain soon! Pan Sen’s new background hero story has already mentioned it. At the end of the "Gun of the Gun" story, I saw the new Protoss. A friend of Pan Sen from another dimension will meet with us soon! so cool! DracCusS: It may be: a) New hero. b) Pan Sen model update. c) Gems reform?} \\
    \midrule
    \textbf{Gold}  & 英雄联盟官方发布星空主题预告短片，或将推出全新巨神峰英雄 
    
    \textit{League of Legends officially released a starry sky theme trailer, or will launch a new Titan Heroes} \\
    \midrule
    \textbf{Pointer Generator}    & LOL官方\underline{发布了一个短片视频短片内容为，其短片内容为潘森正在凝视夜} \underline{空中被星云所围绕亮光} （组图）
    
    LOL officially \underline{released a short video , its short video content is that Pan Sen is } \underline{gazing at the night sky surrounded by nebula light} (Pictures) \\
    \midrule
    \textbf{Transformer-abs} & 英雄联盟官方发布新短片，\uwave{称将于明年1月发行}；英雄联盟新短片显示，\uwave{新将于9月下旬发布}。（组图） 
    
    \textit{The League of Legends officially released a new short film, \uwave{said it will be released in January next year}. The new League of Legends video shows that \uwave{new will be released in late September}. (Pictures)}
    \\
    \midrule
    \textbf{BERT-abs} & 英雄联盟公布新英雄预告：巨神峰新英雄是星灵来自银河？潘森新英雄将加入峡谷，宝石重做巨神之枪（组图） 
    
    \textit{League of Legends announced the new hero trailer: Is the new hero of Titan Peak from the Milky Way? Pan Sen’s new hero will join the canyon, and the gem will be remade the Titan’s Spear (Pictures)}
    \\
    \bottomrule
    \end{tabular}%
\end{table}%
\end{CJK*}

\subsection{Results}
Since the original summarization metric ROUGE~\cite{lin2004rouge} is made only for English, we follow the method of~\cite{hu2015lcsts} and map the Chinese words to numbers. Specifically, the Chinese text is split by characters, and the English words and numbers will be split by space. 
\begin{CJK*}{UTF8}{gkai}
For example, ``Surface Phone将装载Windows 10 (\textit{The Surface Phone will be loaded with Windows 10})" will be transformed to 
``surface/phone/将/装/载/windows/10" and then mapped to numeral IDs.
\end{CJK*}

As shown in Table \ref{tab:results}, the abstractive models have better results on \method test set, which is consistent with our analysis in Section \ref{sec:analysis}. 
The simple abstractive baseline, pointer generator, has performed better than BERT-based extractive models, which means that extractive methods have many performance limitations in \method. 

We further evaluate abstractive models based on adequacy and deducibility level. The results shown in Table \ref{tab:adres} indicate that this model performs well on examples with A=1 where all necessary information can be easily found in the source document. However, on examples that ask for simple deducing or external knowledge, the performance degrades significantly.



\subsection{Case study}
We illustrate the differences between abstractive models with a typical example in Table \ref{tab:case}. As stated in previous work~\cite{see2017get,zhang2018abstractiveness}, the pointer generator tends to copy directly from the original document instead of generating from vocabulary, which makes the output less abstractive. Besides, although it has used the coverage mechanism to avoid repetition, it still suffers the most from meaningless duplication. For Transformer-based models, the random initialized model Transformer-abs introduces fake information, while the BERT-abs performs much better in both capturing important information and generating fluent summaries.

\section{Conclusion}
\label{sec:conclusion}
We present CNewSum, a high-quality summarization dataset composed of human-written summaries to fill up the lack of news summarization dataset in Chinese. 
We annotate all test set with adequacy and deducibility scores to help abstractive models figure out how to generate a more human-friendly summary. Finally, we report results of several popular extractive and abstractive baselines on the dataset for future research.

\section*{Acknowledgments}
The authors would like to thank Huiying Lin and many language annotators for help on preparing the data. Lei Li is not supported by any funding during this project. 



\bibliographystyle{unsrt}
\bibliography{paper}

\end{document}